\documentclass[10pt, logo, copyright]{nvidiatechreport}

\usepackage{mdframed}
\usepackage[utf8]{inputenc} % allow utf-8 input
\usepackage[T1]{fontenc}    % use 8-bit T1 fonts

\usepackage{amsfonts}       % blackboard math symbols
\usepackage{nicefrac}       % compact symbols for 1/2, etc.
\usepackage{microtype}      % microtypography
\usepackage[dvipsnames]{xcolor}         % colors
\usepackage{multirow}
\usepackage{multicol}
\usepackage{graphicx}
\usepackage[round]{natbib}
\usepackage{tabto}
\usepackage{xspace}
\usepackage{amsmath}
\usepackage{adjustbox}
\usepackage{enumitem}
\usepackage{wrapfig}
\usepackage{dblfloatfix}

\usepackage{footmisc}

\usepackage{float}
\usepackage{siunitx}
\usepackage{enumitem}

\usepackage{natbib}
\newcommand{\dataset}{ProCUA-SFT\xspace}

\usepackage{hyperref}
\usepackage{url}
\usepackage{booktabs}
\usepackage{multirow}
\usepackage{enumitem}
\usepackage{adjustbox}
\usepackage{tabularx}
\usepackage{arydshln}
\usepackage{wrapfig}

\usepackage{multirow}
\usepackage{colortbl}
\usepackage{amsmath}
\usepackage{makecell}
\usepackage{hhline}
\usepackage{array}
\usepackage{diagbox}
\usepackage{graphicx}
\usepackage{adjustbox}
\usepackage{comment}

\usepackage{fp} % Required for floating-point calculations

\usepackage{xcolor}

\title{\dataset Technical Report}

\author{%
  Jaehun Jung\textsuperscript{*,1,2},
  Ximing Lu\textsuperscript{*,1,2},
  Brandon Cui\textsuperscript{*,1},
  Muhammad Khalifa\textsuperscript{1},
  Shaokun Zhang\textsuperscript{1},
  Hao Zhang\textsuperscript{1},
  Jin Xu\textsuperscript{1},
  Amala Sanjay Deshmukh\textsuperscript{1},
  Karan Sapra\textsuperscript{1},
  Andrew Tao\textsuperscript{1},
  Yejin Choi\textsuperscript{1},
  Jan Kautz\textsuperscript{1},
  Mingjie Liu\textsuperscript{*,1,\textdagger},
  Yi Dong\textsuperscript{1,\textdagger}%
}

\makeatletter
\newcommand\blfootnote[1]{{\renewcommand\thefootnote{}\footnotetext{#1}}\addtocounter{footnote}{-1}}
\makeatother

\begin{abstract}
Training computer-use agents (CUAs)---models that interact with graphical
desktops through screenshots and keyboard/mouse actions---requires large-scale,
diverse trajectory data collected in full desktop environments.
The largest public resource, AgentNet (22.5K human trajectories), leads to
\emph{negative} transfer when used for supervised fine-tuning (SFT): continuing
training UI-TARS~7B on AgentNet causes OSWorld success rate to fall from 26.3\%
to 8--10\%.
We present \textbf{ProCUA-SFT}, a dataset of \textbf{3.1M} step-level SFT
samples distilled from 93K synthetic trajectories across 2,484 application
combinations.
The dataset is produced by a fully automated pipeline that
(i)~synthesizes \emph{grounded} tasks on live desktops seeded with real-world
content---912 spreadsheets from SpreadsheetBench, ${\sim}$10K
permissively-licensed presentations from Zenodo10K, and multi-application
OSWorld configs---and
(ii)~verifies each task's feasibility through binary precondition checking
before rollout.
A single VLM (Kimi-K2.5) serves as goal generator, precondition judge, and
trajectory executor, eliminating planner--actor capability gaps.
Each trajectory is expanded into step-prefix samples that exactly reproduce the
context layout seen at inference time.
Fine-tuning UI-TARS~7B on ProCUA-SFT for one epoch yields \textbf{45.0\%} on
OSWorld---an 18.7 percentage-point improvement over the base model and over
35\% above AgentNet-trained counterparts. 
A subset of ProCUA was incorporated into the training data for the Nemotron 3 Nano Omni model~\citep{Deshmukh2026Nemotron3N}, contributing to its computer-use capabilities.
\vspace{2mm}
\newline
\textbf{Data on Hugging Face:} \href{https://huggingface.co/datasets/nvidia/ProCUA-SFT}{ProCUA-SFT}
\end{abstract}

\begin{document}
\maketitle
\blfootnote{\textsuperscript{1}NVIDIA\quad\textsuperscript{2}University of Washington}
\blfootnote{\textsuperscript{*}Equal contribution\quad\textsuperscript{\textdagger}Corresponding authors}

\section{Introduction}
\begin{comment}
We release \dataset, a dataset of 3.1M SFT samples for ...

Main narrative\\
- Not much of off-the-shelf datasets out there, so we release this \\
- In particular AgentNet \citep{opencua}, the only viable public resource, leads to poor scaling - we significantly improve both performance (across scale) \& intrinsic data quality (e.g., task diversity, difficulty) \\

* Also an implication that we will keep iterating on the data, this is the first version \\
\end{comment}
\label{sec:intro}

Computer-use agents (CUAs) perceive graphical desktops through screenshots and
act through keyboard and mouse commands.
Yet progress remains early: open-source CUAs such as UI-TARS-7B~\citep{qin2025uitars} and OpenCUA-7B~\citep{wang2025opencua} have reached only 26.3\% and 25.0\% on OSWorld~\citep{xie2024osworld}, a benchmark of realistic desktop tasks, and further progress is bottlenecked by training data.
Unlike text-only instruction tuning, each CUA trajectory requires a fully
instantiated desktop---booted VM, installed applications, realistic file
content---making large-scale collection substantially more expensive than
curating text corpora.

The largest public CUA training resource, AgentNet~\citep{wang2025opencua},
provides 22.5K human-annotated trajectories spanning three operating systems.
Despite its scale and scope, fine-tuning UI-TARS~7B on AgentNet for one epoch
causes OSWorld success rate to fall from 26.3\% to 8--10\%
(Figure~\ref{fig:cua-training-runs}).
We attribute this negative transfer to three factors: limited task diversity
(the majority of trajectories are single-application workflows with a median of
17 steps), the absence of complex cross-application reasoning tasks, and
annotation noise inherent in crowd-sourced demonstrations.
These observations motivate a fully synthetic approach---but na\"{i}ve synthetic
generation faces its own challenge: LLM-generated tasks often reference
non-existent files or unavailable applications, wasting rollout compute on
infeasible goals.

We introduce \textbf{ProCUA-SFT}, a dataset of 3.1M step-level SFT samples
from 93K synthetic trajectories spanning 2,484 application combinations.
The pipeline is built around a single VLM---Kimi-K2.5~\citep{kimiteam2026kimik25}---that
serves as goal generator, precondition judge, and trajectory executor.
Four design decisions define the pipeline:

\begin{enumerate}[leftmargin=*,nosep]
\item \textbf{Grounded task synthesis with precondition verification.}
  The VLM generates each task goal together with a set of binary preconditions
  (e.g., ``Does the file \texttt{Q3.xlsx} exist on the Desktop?'').
  A separate judge prompt independently verifies each precondition against the
  current desktop screenshot and OS-level config; only goals whose
  preconditions all pass proceed to rollout, with failed verdicts fed back for
  iterative retry~(\S\ref{sec:method-goal}).

\item \textbf{Real-world content seeding.}
  Desktops are initialized with 912 real-world spreadsheets from
  SpreadsheetBench~\citep{ma2024spreadsheetbench} (tables exceeding 100
  columns and 20K rows), ${\sim}$10K permissively-licensed presentations from
  Zenodo10K~\citep{forceless2024zenodo}, and multi-application configs from
  OSWorld~\citep{xie2024osworld}.
  This yields tasks that operate on genuine complex documents---formulas,
  charts, multi-sheet references---rather than empty or default application
  states~(\S\ref{sec:method-sources}).

\item \textbf{Single-VLM rollout.}
  Using the same model for goal synthesis and trajectory execution closes the
  planner--actor capability gap: the model never proposes goals beyond what it
  can carry out, which maximizes the fraction of trajectories that reach
  successful completion~(\S\ref{sec:method-rollout}).

\item \textbf{Step-prefix expansion.}
  Each trajectory of $T$ steps yields $T$ training samples; sample~$t$ contains
  the full screenshot--action history up to step~$t$, reproducing the
  growing-context layout the model encounters at inference
  time~(\S\ref{sec:method-sft}).
\end{enumerate}

\noindent
The pipeline runs on decoupled infrastructure---local KVM via Singularity and
serverless NVCF---enabling parallel collection across heterogeneous
compute~(\S\ref{sec:method-infra}).
After one epoch of SFT on UI-TARS~7B, ProCUA-SFT achieves \textbf{45.0\%} on
OSWorld, an 18.7~pp gain over the base model and more than 35~pp above
AgentNet-trained models (Figure~\ref{fig:cua-training-runs}). As evidence of its utility, a portion of ProCUA was incorporated into the training data for the Nemotron 3 Nano Omni model~\citep{Deshmukh2026Nemotron3N}, contributing to that model's computer-use capabilities.

\begin{comment}
Beyond the dataset itself, we make two additional contributions.
First, a \textbf{trajectory profiling framework} that models each trajectory as
a directed screen-transition graph and extracts 18 topology
metrics---linearity, cycle count, app-switch frequency, revisit ratio,
etc.---for quantifying structural complexity~(\S\ref{sec:analysis}).
Second, a \textbf{complexity-aware curation pipeline} that identifies rare
application combinations ($\leq$3 corpus occurrences $\to$ 2,065
trajectories), uses their goals and summaries as few-shot seeds to synthesize
tasks for underrepresented workflows, and selects diverse training subsets via
round-robin sampling over 2,484 app-combination
buckets~(\S\ref{sec:curation}).
All data and code are publicly released.
\end{comment}

%\section{Related Works}
\section{Related Work}
\label{sec:related}

\paragraph{Vision-based GUI agents.}
Early GUI agents relied on structured inputs---DOM trees, accessibility
APIs---to perceive and act on
interfaces~\citep{shi2017wob,liu2018miniwob,deng2023mind2web}.
Screenshot-based models remove this dependency.
Pix2Act~\citep{shaw2023pix2act} trains a Pix2Struct encoder--decoder with
demonstrations and tree-search self-play on MiniWoB++ and web tasks.
CogAgent~\citep{hong2023cogagent} introduces dual-resolution visual encoding
(1120$\times$1120) for recognizing small UI elements.
SeeClick~\citep{cheng2024seeclick} and Ferret-UI~\citep{you2024ferretui}
advance GUI grounding---through grounding pre-training and resolution-adaptive
encoding for mobile UIs, respectively---while OS-Atlas~\citep{wu2024osatlas}
and UGround~\citep{gou2024uground} scale grounding corpora to 13M+ and 10M
elements across platforms.
At the full-system level, UI-TARS~\citep{qin2025uitars} combines enhanced
perception with system-2 reasoning to reach state-of-the-art on 10+
benchmarks; ShowUI~\citep{lin2024showui} reduces visual token cost via
UI-guided selection; Aguvis~\citep{xu2024aguvis} trains a pure-vision agent
across web, mobile, and desktop; and
ScribeAgent~\citep{shen2024scribeagent} converts screenshots to structured
text, bypassing vision entirely.
ProCUA-SFT is a training-data contribution complementary to all of these
architectures.

\paragraph{Benchmarks.}
OSWorld~\citep{xie2024osworld} provides 369 tasks in real
Linux/Windows/macOS VMs with unrestricted keyboard/mouse control and is our
primary evaluation target.
WebArena~\citep{zhou2024webarena} (812 tasks) and
VisualWebArena~\citep{koh2024visualwebarena} (910 tasks) evaluate web agents
on self-hosted sites.
Windows Agent Arena~\citep{bonatti2024windowsarena} provides 150+ Windows
desktop tasks;
AndroidWorld~\citep{rawles2024androidworld} offers 116 programmatic tasks
across 20 Android apps; and
MiniWoB++~\citep{liu2018miniwob} provides 100+ simplified web micro-tasks.

\paragraph{Training data for GUI agents.}
Human-demonstration datasets include
AITW~\citep{rawles2023aitw} (715K Android episodes),
Mind2Web~\citep{deng2023mind2web} (2K web tasks with crowdsourced actions),
and AgentNet~\citep{wang2025opencua} (22.5K cross-platform trajectories).
AgentNet is the largest desktop-targeted resource to date, yet as we show
in~\S\ref{sec:experiments}, it causes negative transfer when used for SFT.

Recent works synthesize trajectories automatically to bypass the cost of human
annotation.
In web environments, AgentTrek~\citep{xu2024agenttrek} replays web tutorials
to produce 10K trajectories, and
Explorer~\citep{pahuja2025explorer} generates 94K trajectories via exploration
across 49K URLs.
InSTA~\citep{trabucco2025insta} annotates 150K websites and filters by
LLM-judged success;
Go-Browse~\citep{gandhi2025gobrowse} applies Go-Explore--style graph search;
and SynthAgent~\citep{wang2025synthagent} introduces dual refinement of tasks
and trajectories.
In desktop settings, AgentSynth~\citep{xie2025agentsynth} composes subtasks
into long-horizon tasks via information asymmetry, while
PC~Agent~\citep{he2024pcagent} and PC~Agent-E~\citep{he2025pcagente}
augment small human-trajectory seed sets with LLM-synthesized alternatives.
Orthogonally, DigiRL~\citep{bai2024digirl} applies online RL to Android
device control, and AgentRefine~\citep{fu2025agentrefine} refines failed
trajectories through self-correction before SFT.

ProCUA-SFT differs from prior synthesis work in three respects:
(i)~it verifies task feasibility through in-loop binary precondition checking
rather than post-hoc filtering;
(ii)~it seeds desktops with externally sourced complex content (real
spreadsheets, real presentations) rather than default or empty states; and
(iii)~at 3.1M samples from 93K trajectories across 2,484 application
combinations, it is the largest open-source desktop CUA training dataset.

\section{\dataset}
% !TEX root = ../main_tech_report.tex

% TODO: replace with pipeline diagram
\begin{figure}[!htbp]
\centering
\includegraphics[width=0.94\textwidth]{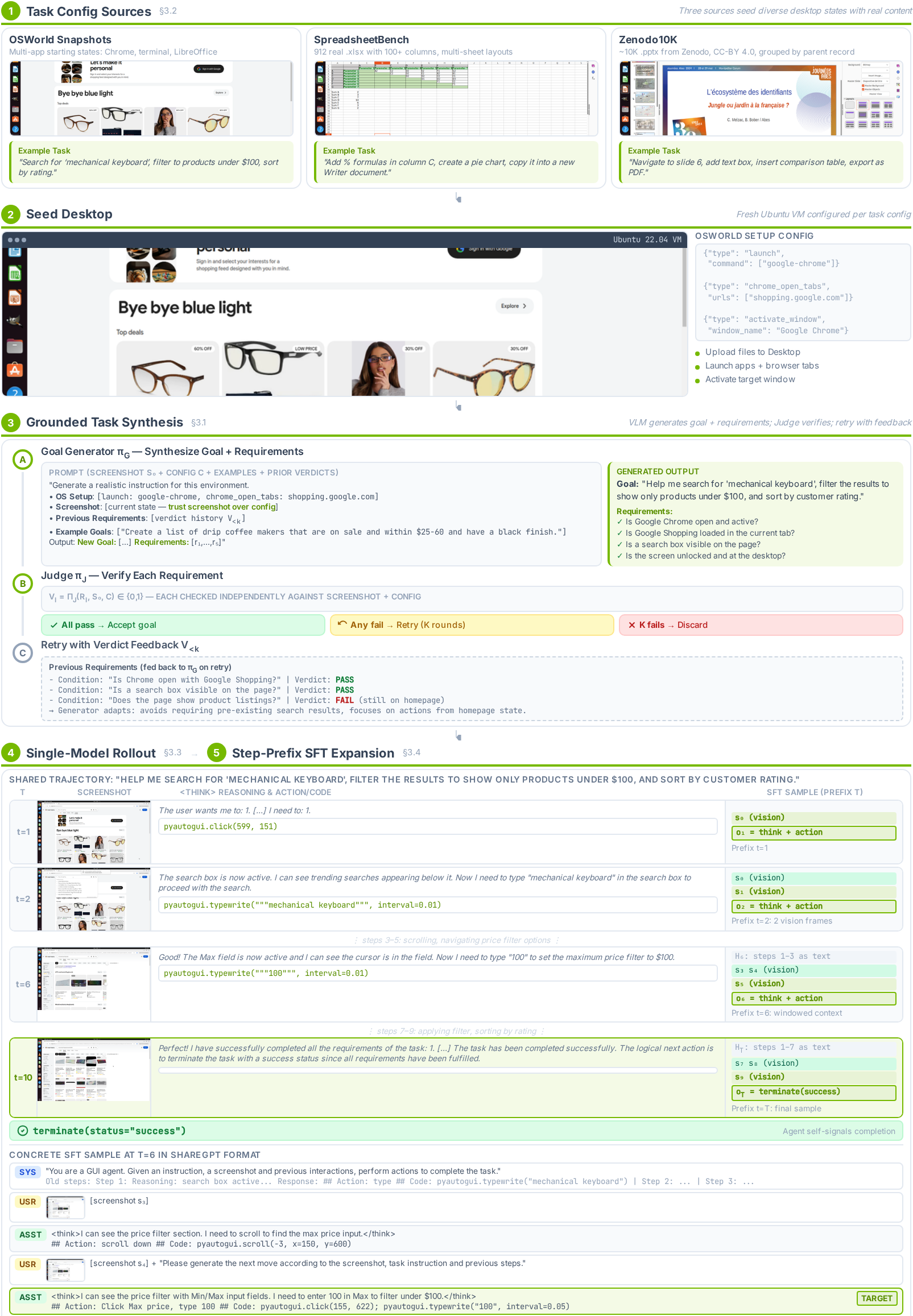}
\vspace{-2.4mm}
\caption{\textbf{\dataset pipeline.} A fresh desktop is seeded from one of three task-config sources (OSWorld, SpreadsheetBench, Zenodo10K). The same VLM then synthesizes a grounded goal from the screenshot and OS state, rolls out a trajectory toward it, and signals success/failure via a reserved \texttt{terminate} call. Each trajectory is expanded into multiple step-prefix SFT samples in ShareGPT format.}
\label{fig:pipeline}
\end{figure}

The goal of \dataset is to provide diverse, grounded, and difficult computer-use trajectories at a scale large enough to drive supervised fine-tuning. Three design choices distinguish the pipeline from a naive ``run a strong agent on a benchmark and harvest its rollouts'' recipe: (i) tasks are \emph{synthesized from the desktop}, with explicit precondition checks, so that the goal is provably grounded in what the agent can actually do (\S\ref{sec:method-goal}); (ii) the desktop is \emph{seeded with real, externally sourced content}---real-world spreadsheets and presentations under permissive licenses---so that hard cross-application reasoning is possible at all (\S\ref{sec:method-sources}); and (iii) the data-collection infrastructure decouples model inference from environment execution and is portable across heterogeneous compute backends, which is what lets the same code scale data collection from either local KVM hosts or serverless cloud VMs (\S\ref{sec:method-infra}). Throughout the pipeline we use a single VLM, Kimi-K2.5~\citep{kimi-k2.5}, for all policy roles---goal generation, requirement verification, and trajectory rollout---under different prompts. Figure~\ref{fig:pipeline} summarizes the resulting end-to-end flow. %\mk{writing is a bit dense here. Also not explained why we choose Kimi in particular.}

\subsection{Grounded Task Synthesis}
\label{sec:method-goal}

A common failure mode of synthetic CUA data is the \emph{infeasible task}: the generator asks the agent to ``open the Q3 report on the Desktop'' when no such file exists, or to ``send the email draft'' when no mail client is configured. Such tasks waste rollout budget and---worse---teach the SFT model to invent state. Conversely, restricting tasks to whatever is visible on the screen is too conservative: it misses files placed by the setup config, browser tabs not yet focused, and applications launchable from the system menu but not currently in the foreground.

We address this by treating task synthesis as a \emph{grounded generation} problem with an in-loop verifier. Let $\mathcal{S}$ denote the space of desktop states, and let $\pi_G$ and $\pi_J$ denote a goal-generator policy and a Judge policy; in our implementation both are realized by the same VLM with different prompts.

\paragraph{Generating goals with verifiable requirements.} Rather than asking the generator for a goal in isolation, we ask it to emit a goal \emph{together with} a small set of binary preconditions that must hold in the current desktop for the goal to be achievable:
\begin{equation}
\pi_G : (s_0,\; c,\; a,\; V_{<k}) \;\longmapsto\; \big(g^{(k)},\; R^{(k)}\big),
\qquad R^{(k)} = \{r_1, r_2, \ldots, r_5\},
\label{eq:goal-gen}
\end{equation}
where $s_0 \in \mathcal{S}$ is the initial screenshot; $c$ is the structured OSWorld setup config (files placed on disk, applications launched, browser tabs opened, shell commands run); $a$ is a style-anchor instruction sampled from AgentNet~\citep{opencua}; $V_{<k}$ is the set of requirement--verdict pairs accumulated from prior failed attempts on the same VM (with $V_{<1} = \emptyset$); and each requirement $r_i : \mathcal{S} \to \{0, 1\}$ is an objective, binary predicate on desktop state. For example, alongside a goal of \emph{``open the quarterly report and add a chart of the revenue column,''} the generator might emit requirements such as ``Does \texttt{Q3.xlsx} exist on the Desktop?'', ``Is LibreOffice Calc installed?'', and ``Is the screen unlocked and at the desktop?''. These precondition questions are precisely what the Judge will subsequently adjudicate.

\paragraph{Verifying with the Judge and verdict-conditioned retry.} Each requirement is independently evaluated by the Judge policy against the current observation, producing a yes/no verdict
\begin{equation}
v_i^{(k)} \;=\; \pi_J(r_i,\, s_0,\, c) \;\in\; \{0, 1\}.
\label{eq:judge}
\end{equation}
We \emph{accept} the candidate goal iff $\bigwedge_i v_i^{(k)} = 1$. Otherwise we update $V_{<k+1} = V_{<k} \cup \{(r_i, v_i^{(k)})\}_i$ and re-invoke $\pi_G$, up to a maximum of $K$ rounds. The verdict feedback carried in $V_{<k}$ is what steers subsequent attempts away from preconditions already proven false on this VM and toward conditions the environment can actually satisfy.

Two properties make this loop a grounding mechanism rather than self-discipline scaffolding. First, requiring $\pi_G$ to commit to binary, verifiable $r_i$ \emph{alongside} the goal yields a hand-off that $\pi_J$ can actually adjudicate---fluent but ungrounded preconditions are exactly what the verifier rejects. Second, because $c$ is supplied to both $\pi_G$ and $\pi_J$, the loop closes over resources promised by the setup but not visible at $t{=}0$ (a file deposited on disk by an \texttt{upload\_file} step, a shell-launched server that has not yet rendered a window), broadening task diversity beyond surface visual content. We run the verify-and-retry loop ($K > 1$) on a portion of the released data and a single-shot variant ($K = 1$, no Judge call) on the remainder; both share the formulation in Eq.~\eqref{eq:goal-gen}.

\subsection{Sourcing Complex, Permissively-Licensed Content}
\label{sec:method-sources}

The hardest CUA tasks---reading a number from a spreadsheet, looking it up on the web, summarizing the result into a slide deck---are bottlenecked not by the agent's action space but by the \emph{content} on the desktop. Empty Calc workbooks and stock template presentations do not support such workflows. We therefore seed each trajectory from one of three task-config sources, two of which inject genuinely complex external content:

\begin{itemize}[leftmargin=*]
    \item \textbf{Vanilla OSWorld snapshots.} OSWorld \citep{osworld} ships a set of example setup configs that, after we exclude entries depending on Google Drive credentials, give us multi-application starting states covering web browsing, terminal sessions, and the LibreOffice suite. We use these for breadth.
    \item \textbf{SpreadsheetBench.} SpreadsheetBench \citep{spreadsheetbench} is a collection of 912 real-world spreadsheets harvested from online Excel forums, with tables that exceed 100 columns and 20{,}000 rows and that contain non-standard layouts, multiple sheets, and rich non-textual content. We adapt only the workbooks (not the benchmark's reference solutions or test cases) into setup configs that upload the \texttt{.xlsx} to the Desktop and open it in LibreOffice Calc. The benchmark releases the workbooks under a research license; we use them only to instantiate desktop state and do not redistribute the files themselves.
    \item \textbf{Zenodo10K presentations.} Zenodo10K \citep{zenodo10k} is a public collection of $\sim$10{,}000 \texttt{.pptx} files mined from the Zenodo open-research repository. We use the subset released under a permissive reuse license (CC-BY 4.0). Beyond its raw scale, the dataset preserves Zenodo's parent-record IDs, which group co-published artifacts---a property we exploit to seed multi-file desktop states (described below).
\end{itemize}

\paragraph{Seeding multi-file desktops via Zenodo grouping.} Real users rarely keep a single isolated document on their desktop: drafts, supplements, and related decks tend to sit alongside each other, and a class of useful tasks (compare two versions, fold an appendix into a main deck, locate a file by name in the file manager) only makes sense in that multi-file setting. Sampling presentations independently would underspecify this startup state. We therefore sample one Zenodo parent record per trajectory, upload all of its member files to the Desktop, and open between one and three of them in LibreOffice Impress (with weights $0.5/0.3/0.2$); the unopened files remain as bystander documents that the agent can discover via the file manager.

The combination of grounded goal synthesis (\S\ref{sec:method-goal}) and these external seed corpora is what makes the resulting tasks \emph{simultaneously} hard and feasible: the generator is asked to invent a goal that exercises the spreadsheet or presentation that has been opened in front of it, but is constrained by binary requirements to only ask for things the desktop can actually deliver. Per-source goal-synthesis prompts further specialize this objective---e.g., the SpreadsheetBench prompt requires the goal to reference visible columns, sheet names, and data patterns and to chain $2{-}3$ sub-tasks (cross-sheet aggregation, charting, export to Writer/CSV), while a multi-app variant of the Zenodo prompt encourages workflows that span Impress, Calc, Writer, Chrome, the file manager, and the terminal.

\subsection{Single-Model Rollouts and Self-Termination}
\label{sec:method-rollout}

Once a goal is synthesized, the same VLM rolls out a trajectory toward it. Using a single model for both goal synthesis and acting is deliberate: the generator's notion of ``what is achievable'' coincides with the actor's actual capability, which prunes the gap in which a stronger planner proposes goals that a weaker actor cannot execute. At each step the model receives the current screenshot, the goal, and a windowed history (see below), and produces a chain of thought followed by an \textbf{Action}/\textbf{Code} pair. The code is either raw \texttt{pyautogui} (\texttt{click}, \texttt{moveTo}, \texttt{hotkey}, \texttt{typewrite}, \texttt{scroll}, \texttt{dragTo}, \dots) over absolute screen coordinates, or one of two reserved control functions: \texttt{wait()} for installs and slow UI loads, and \texttt{terminate(status, answer)}, with which the agent self-signals success or failure and optionally returns a textual answer. We rely on \texttt{terminate} rather than an external success classifier: each rollout therefore carries its own coarse-grained label, which downstream filtering can use as a soft quality signal without requiring per-task evaluators.

\paragraph{Context windowing matched to the SFT format.} High-resolution screenshots make every additional vision frame expensive. We keep the three most recent screenshots in vision form and convert older steps into a short textual summary block (``\texttt{Old steps: Step $k$: Reasoning: $\dots$ Response: $\dots$}'') prepended to the system message. Crucially, this exact windowing scheme is reproduced during SFT conversion, so the supervised model is trained on the same context layout it will encounter at inference time---there is no train/test mismatch in how prior steps are presented.

\subsection{From Trajectories to SFT Samples}
\label{sec:method-sft}

A naive single-sample-per-trajectory conversion wastes most of the supervision in a long rollout. We instead apply \emph{step-prefix expansion}. Let a trajectory $\tau$ consist of a goal $g$, an initial screenshot $s_0$, and a sequence of step pairs $\{(o_t, s_t)\}_{t=1}^{T}$, where $o_t = (\text{think}_t, \text{action}_t)$ is the model's output at step $t$ and $s_t$ is the resulting screenshot. We expand $\tau$ into the supervised set
\begin{equation}
\mathcal{D}(\tau) \;=\; \big\{(x_t,\, o_t)\big\}_{t=1}^{T},
\qquad
x_t \;=\; \big(g,\; V_t(\tau),\; H_t(\tau)\big),
\label{eq:step-prefix}
\end{equation}
where $V_t(\tau)$ retains the at-most-three most recent screenshots from $\{s_0, \dots, s_{t-1}\}$ in vision form and $H_t(\tau)$ summarizes the older $\{o_i\}_{i \le t-4}$ as a textual block. Each prefix length therefore contributes its own training sample, exposing the SFT model to every state the actor saw during rollout---including very early states, where most teaching about how to \emph{begin} a task is concentrated---rather than only the final state of completed runs.

Each sample is emitted in the LLaMA-Factory ShareGPT format with images interleaved into the conversation via \texttt{<image>} placeholders and the corresponding paths listed in a parallel \texttt{images} field. The supervision target is the chain of thought wrapped in \texttt{<think>\dots</think>}, followed by the \texttt{\#\# Action:} / \texttt{\#\# Code:} block; the user-facing turns reproduce the same system prompt, optional ``Old steps:\dots'' summary (active only when $t > 3$), and instruction template used at rollout time. We discard any sample whose referenced screenshots are missing, then shuffle and shard the remaining records.

\subsection{Scaling Synthetic CUA Collection}
\label{sec:method-infra}

Generating millions of steps requires running thousands of full Linux desktops in parallel, which is the practical bottleneck of CUA data collection rather than model inference. Two architectural decisions made this tractable on heterogeneous compute.

\paragraph{Decoupled inference and environment.} The VLM is served as a stateless OpenAI-compatible vLLM endpoint (Ray-based, tensor-parallel across the model's GPUs); each data-collector process owns its own VM and talks to the endpoint over HTTP. Because the actor is stateless and the environment is the slow side, we can scale model and environment fleets independently. Within a single Slurm reservation we colocate the vLLM Ray cluster and the collectors on the same GPU nodes, using the otherwise-idle CPU cores on those nodes to host VMs; this avoids paying inter-node bandwidth for the dominant traffic of high-resolution screenshots.

\paragraph{Pluggable VM backend.} We implement two interchangeable VM backends behind a common interface:
\begin{itemize}[leftmargin=*]
    \item \textbf{Singularity (local).} An OSWorld QEMU/KVM stack packaged as a Singularity (Apptainer) \texttt{.sif} image that runs without root and without a Docker daemon, which makes it deployable on standard HPC nodes. When the host exposes \texttt{/dev/kvm} the VM uses hardware acceleration; otherwise the same image transparently falls back to software emulation, trading throughput for portability.
    \item \textbf{NVCF (serverless).} For pools where local KVM is unavailable or local capacity is saturated, the same collector code drives NVIDIA Cloud Functions (NVCF; \citealp{nvcf}), NVIDIA's serverless platform for GPU-attached workloads. NVCF handles auto-deploy, auto-scale, and auto-undeploy of the OSWorld container; we expose it to the rest of the pipeline through a small local HTTPS-to-HTTP proxy that forwards the OSWorld API, Chrome DevTools, and VLC endpoints. To avoid burning paid GPU time on a deploy that will fail anyway, we pre-download all setup files to a local cache \emph{before} committing the NVCF deploy, and only proceed if every download succeeds.
\end{itemize}

\paragraph{Throughput-oriented orchestration.} Each collector runs an asyncio thread pool with two stages---\emph{init} (sample a config, boot the VM, apply the setup) and \emph{collect} (synthesize a goal, roll out actions)---so that a slow boot never blocks an in-progress rollout. A semaphore caps concurrent VMs per node; sequential boot delays prevent the thundering-herd that otherwise destabilizes the QEMU/KVM stack when many guests start at once. Trajectories are checkpointed to disk after every step so that a node-level failure costs at most one in-flight rollout. \textbf{[total trajectories / num resources (how many VMs used in parallel)]}

\section{Empirical Validation}
\label{sec:experiments}
% - Comparison with AgentNet, scaling curve across steps (both on 1 epoch run) \\

\begin{figure*}[!t]
\includegraphics[width=\textwidth]{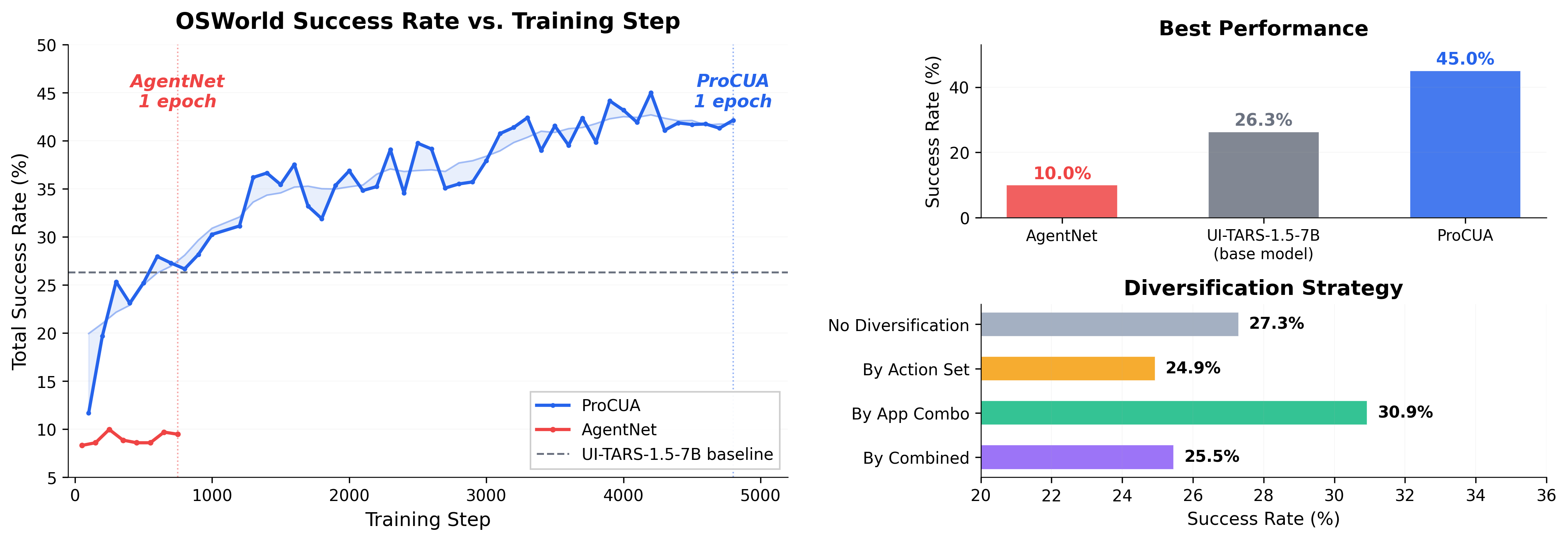}
%\caption{OSWorld evaluation curves as we continue training UI-Tars 1.7 7B on Agent Net and ProCUA for 1 epoch. After 1 epoch of training on Agent Net, degradation on OSWorld; whereas, after 1 epoch of training with ProCUA, we are able to improve UI-Tars to up to a 45\% success rate.}
  \caption{%
    \textbf{(a)}~OSWorld success rate during SFT on UI-TARS~7B.
    ProCUA (blue) rises steadily from 11.7\% to a peak of 45.0\% over
    ${\sim}$4{,}800 steps, while AgentNet (red) plateaus at 8--10\% after
    one epoch (750 steps)---below the pretrained UI-TARS-1.5-7B baseline
    of 26.3\% (dashed).
    The light band shows the trend after moving-average smoothing.
    \textbf{(b)}~Best checkpoint comparison: ProCUA reaches 45.0\%,
    outperforming both the base model (26.3\%) and AgentNet (10.0\%) by
    18.7~pp and 35.0~pp, respectively.
    \textbf{(c)}~Ablation over data diversification strategies at a matched
    training budget (${\sim}$700 steps).
    Round-robin sampling by app combination (``By App Combo'') achieves
    30.9\%, the only strategy to exceed the base-model baseline; selection
    by action set or the combined nested strategy underperforms the
    non-diversified variant (27.3\%).
  }
\label{fig:cua-training-runs}
\end{figure*}

\begin{figure}[!t]
    \centering
    \includegraphics[width=0.99\linewidth]{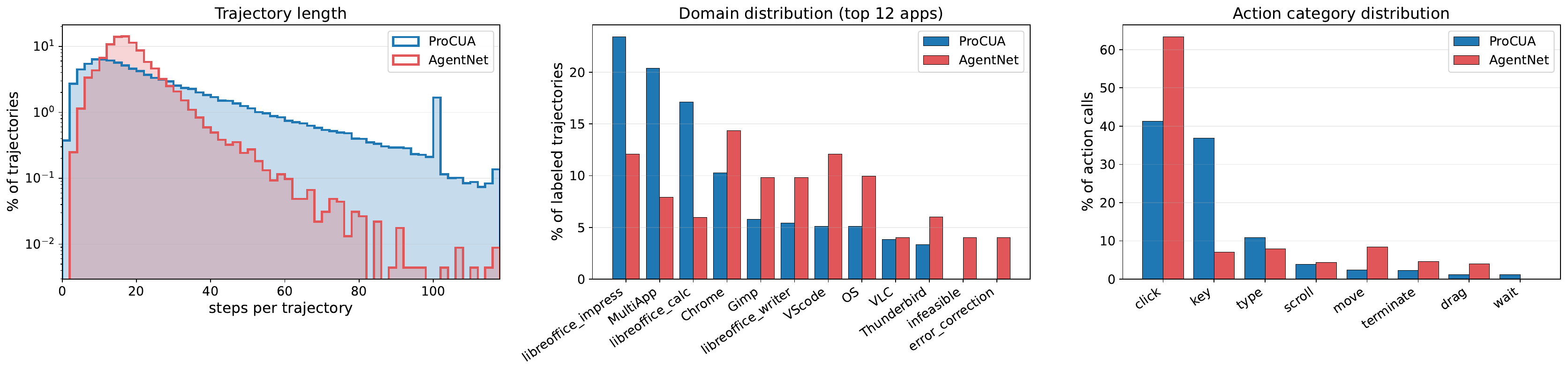}
      \caption{\textbf{Comparison of dataset statistics for ProCUA and AgentNet.}
  \emph{Left:} Distribution of trajectory lengths (steps per trajectory), with
  the $y$-axis on a log scale. ProCUA trajectories are markedly longer-horizon ($\mu \approx 29.7$ vs.\
  $18.6$ steps) with a substantially heavier tail. \emph{Center:} Domain
  distribution across the twelve most frequent applications, as a percentage of
  labeled trajectories. \emph{Right:} Action category distribution, as a
  percentage of all action calls; ProCUA relies more heavily on keyboard
  actions, whereas AgentNet is more click-dominated.}
    \label{fig:procua-agentnet}
\end{figure}

We continue training UI-Tars 1.5 7B on the Agent Net and ProCUA datasets for 1 epoch. We pack samples to a maximum sequence length of 32k, with a batch size of 512, learning rate $2e-5$, using a cosine learning rate scheduler, and weight decay of $0.1$.

After 1 epoch of training, ProCUA reaches 44.99\% success rate on OSWorld, a 18.68\% improvement over the original UI-Tars performance. On the other hand, AgentNet only leads to degradation from the original model, plateauing at \textasciitilde 8-10\%. Notably, at a similar number of training steps, the ProCUA dataset is able to recover the performance of the original UI-Tars performance. 

Comparing ProCUA to AgentNet, we can identify several factors that likely contribute to ProCUA's  stronger OSWorld performance over AgentNet. First, ProCua trajectories are markedly longer-horizon: their mean length is roughly 1.8X that of AgentNet's (\textasciitilde30 vs \textasciitilde17 steps) and their tail extends much farther, as shown in Figure~\ref{fig:procua-agentnet} (left). Second, ProCUA's app coverage is more aligned with OSWorld's evaluation distribution. The middle panel of Figure~\ref{fig:procua-agentnet} shows that nearly half of all labeled ProCUA trajectories target LibreOffice applications (Impress 24\%, Calc 17\%, Writer 6\%), and another 20\% involve multi-application workflows---both categories that dominate OSWorld. Note that the entire Windows/Mac portion of AgentNet (18k of 22.6k trajectories) omits app-level metadata in the released data, so this comparison is restricted to AgentNet's 5k labeled (Ubuntu) trajectories.

Third, the action mix differs in a way that favors more deterministic interaction. As shown in Figure~\ref{fig:procua-agentnet} (right), AgentNet trajectories are heavily click-dominated (\textasciitilde 63\% of all calls vs.\ \textasciitilde 41\% for ProCUA), whereas ProCUA shifts a much larger fraction of its actions onto keyboard primitives and text entry. Keyboard shortcuts and direct text input are inherently less brittle than pixel-accurate clicks, especially across the kind of fine-grained menu navigation and form-filling that OSWorld tasks demand.

\paragraph{Diversification ablation.}
Since CUA training data is distributed unevenly across application combinations and
action types, we ask: \emph{which axis of diversity matters most for
downstream performance, and what is the best strategy to select a diverse
subset from a large trajectory pool?}
We investigate this by fixing the training budget at ${\sim}$700 steps and
comparing four data-selection strategies applied to the same
111{,}862-trajectory pool (Figure~\ref{fig:cua-training-runs}c).
Three strategies use round-robin sampling---by action set, by app combination,
and a nested combined strategy that
round-robins over app-combination buckets and within each bucket over
action-set sub-buckets, and no diversity-aware selection.
%and a nested combination of both (\S\ref{sec:curation})---each selecting 10K
trajectories.
%The fourth (``No Diversification'') trains on the randomly sampled ProCUA dataset without any diversity-aware sub-selection, evaluated at the same step count.

Round-robin by app combination achieves 30.9\%, outperforming both the
base-model baseline (26.3\%) and the non-diversified variant (27.3\%) by
4.6~pp and 3.6~pp, respectively.
In contrast, selection by action set (24.9\%) and the nested combined strategy
(25.5\%) both underperform the non-diversified baseline.
This result identifies \emph{application-combination coverage} as the dominant
diversity axis for CUA SFT: ensuring broad representation across app
combinations matters more than balancing action types.
%At full scale (${\sim}$4{,}800 steps), ProCUA---which implicitly covers all 2{,}484 app combinations---reaches 45.0\%, indicating that diversity and scale are complementary.

\begin{comment}
CUA data is unevenly distributed across application combinations and action
types.
To identify \emph{which diversity axis matters most}, we fix the training
budget at ${\sim}$700 steps and compare four selection strategies over the
same 111{,}862-trajectory pool (Figure~\ref{fig:training-curves}c):
round-robin by action set, by app combination, a nested combination of both,
and no diversity-aware selection.
Round-robin by app combination achieves 30.9\%---+3.6~pp over the
non-diversified variant (27.3\%) and +4.6~pp over the base model---while
selection by action set (24.9\%) and the nested strategy (25.5\%) both
underperform.
This identifies \emph{application-combination coverage} as the dominant
diversity axis: broad app-combo representation matters more than balancing
action types.
\end{comment}

\section{Analysis}
% ----- Figure -----
\begin{figure*}[!t]
  \centering
  \includegraphics[width=\textwidth]{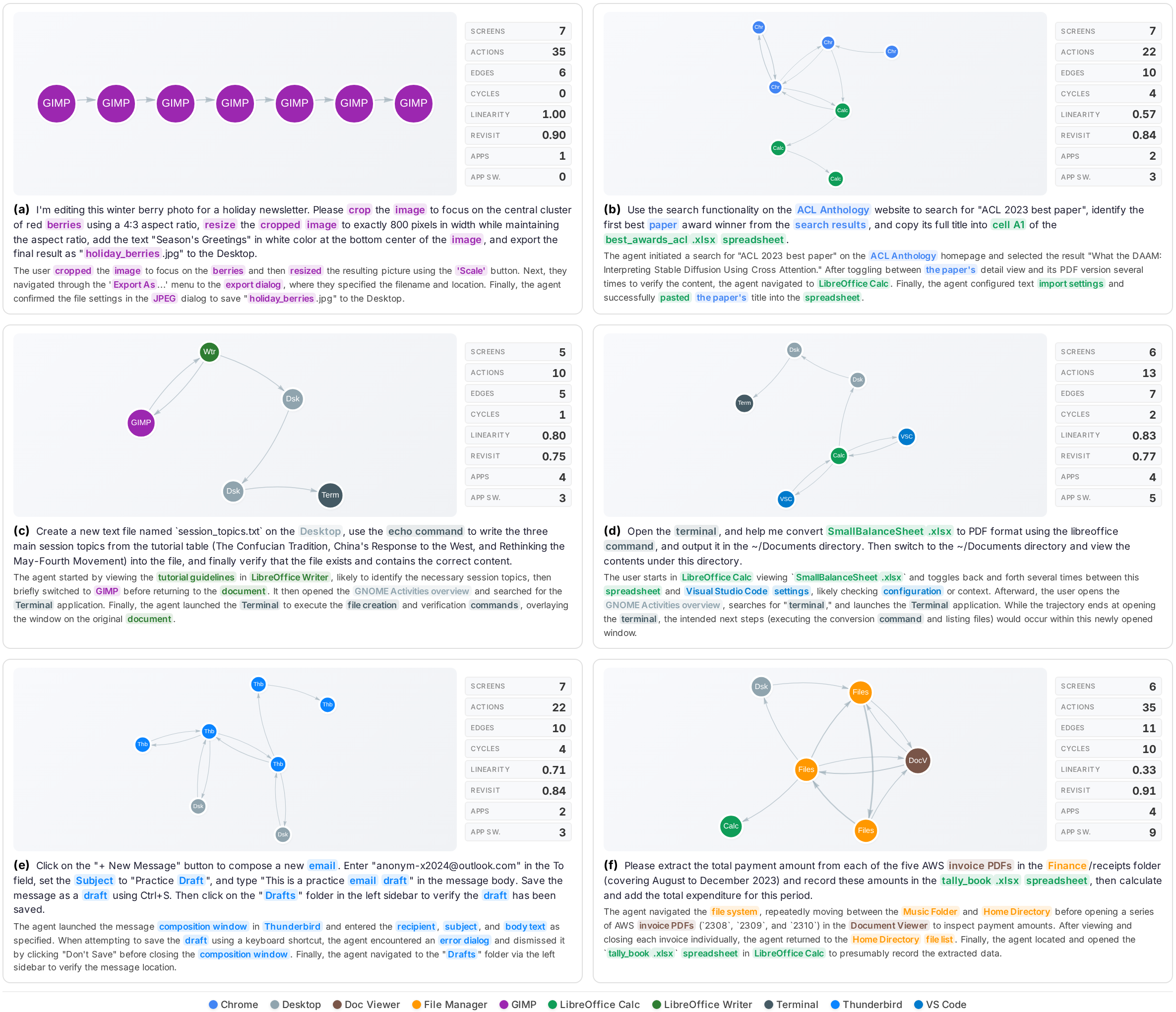}
  \caption{%
    Six representative GUI-agent trajectories illustrating the spectrum of
    structural complexity in our dataset.
    Each panel shows the screen-transition graph (nodes colored by application),
    key graph metrics, the task goal, and a trajectory summary with
    application-related terms color-coded to match their graph nodes.
    \textbf{(a)}~A purely linear, single-app workflow in GIMP
    (linearity\,=\,1.00, 0 cycles).
    \textbf{(b)}~A two-app information-gathering task across Chrome and
    LibreOffice Calc with moderate backtracking (4 cycles, linearity\,=\,0.57).
    \textbf{(c)}~A four-app task with near-linear flow despite involving
    Writer, GIMP, Desktop, and Terminal (linearity\,=\,0.80, 1 cycle).
    \textbf{(d)}~A file-conversion task traversing Calc, VS~Code, Desktop,
    and Terminal with limited branching (linearity\,=\,0.83, 5 app switches).
    \textbf{(e)}~An email-composition task in Thunderbird exhibiting cyclic
    navigation due to error recovery (4 cycles, linearity\,=\,0.71).
    \textbf{(f)}~A highly non-linear, multi-app data-extraction workflow across
    File Manager, Document Viewer, Desktop, and Calc, with the highest
    cycle count (10), lowest linearity (0.33), and most app switches (9)
    among the six examples.
  }
  \label{fig:trajectory-examples}
\end{figure*}

% ----- Analysis section -----
\subsection{Trajectory Graph Analysis}
\label{sec:trajectory-graph-analysis}

To characterize the structural complexity of GUI-agent trajectories, we model
each trajectory as a directed \emph{screen-transition graph}: nodes represent
distinct screens (identified by application and visual state), and edges
represent transitions triggered by agent actions.
From this graph we extract eight metrics that capture complementary aspects of
navigational behavior:
the number of \emph{screens} (nodes) and \emph{transitions} (edges);
the number of \emph{cycles} (circular navigation patterns indicating
backtracking or retries);
total \emph{actions} performed;
the count of distinct \emph{applications} used and the number of
\emph{app switches};
a \emph{linearity score} (fraction of nodes with simple in/out flow; 1.0
denotes a perfectly sequential chain);
and a \emph{screen-revisit ratio} (fraction of screen visits that return to a
previously visited screen).

\paragraph{Dataset-level statistics.}
We profiled \num{34854} trajectories spanning 61 unique application
combinations.
Across a sample of \num{5000} profiled trajectories, the median trajectory
visits 5 screens, traverses 5 transitions, and executes 19 actions.
The median linearity score is 0.86, with 42.3\% of trajectories being
perfectly linear (linearity\,=\,1.0) and only 2.2\% exhibiting highly
non-linear structure (linearity\,$<$\,0.5).
Approximately 35.8\% of trajectories are acyclic, 56.7\% contain 1--5
cycles, and 7.5\% exhibit more than 5 cycles.
Single-application trajectories account for 73.9\% of the dataset (25{,}767
out of 34{,}854), while 18.5\% involve two applications, 6.6\% involve three,
and 0.9\% involve four or more.

\paragraph{Key findings from representative examples.}
Figure~\ref{fig:trajectory-examples} presents six trajectories selected to
illustrate the full complexity spectrum.
Three patterns emerge:

\begin{enumerate}[leftmargin=*,nosep]

\item \textbf{Linear single-app workflows} (panel~a) proceed through a chain
of distinct screens without revisiting any state.
Despite visiting 7 screens and executing 35 actions, the GIMP image-editing
task achieves perfect linearity (1.00) with zero cycles, indicating that the
agent completes each sub-task (crop, resize, export) in a single forward pass.

\item \textbf{Multi-app tasks with moderate branching} (panels~b--e) introduce
cycles when the agent must switch contexts or recover from errors.
In panel~(b), the agent backtracks between a paper detail view and its PDF
version in Chrome before pasting results into Calc (6 backtracks, 4 cycles).
Panel~(e) shows error-recovery branching: the Thunderbird draft-save shortcut
triggers an error dialog, forcing the agent to dismiss and retry
(4 cycles, revisit ratio\,=\,0.84).
Panels~(c) and~(d) involve four applications each but maintain relatively
linear structure (linearity\,$\geq$\,0.80) because the agent moves through
apps in a largely sequential order.

\item \textbf{Highly non-linear multi-app workflows} (panel~f) are characterized
by dense graphs with many cycles.
The invoice-extraction task requires repeatedly navigating between the file
manager, document viewer, and spreadsheet, producing 10 cycles, 13 backtracks,
the lowest linearity (0.33), highest graph density (0.37), and 9 app switches.
Its transition efficiency of 0.17 (unique screens per action) is the lowest
among the six examples, reflecting the high cost of inter-application
coordination.

\end{enumerate}

These examples demonstrate that trajectory complexity is not simply a function
of the number of applications involved: panels~(c) and~(d) involve four apps
yet remain largely linear, while panel~(f) also involves four apps but produces
a highly cyclic graph.
The critical factor is the \emph{pattern of inter-app coordination}---tasks
requiring repeated cross-referencing between applications
(e.g., extracting data from multiple PDFs into a spreadsheet) yield
fundamentally different graph topologies than tasks where applications are
visited sequentially.

% ----- Figure -----
\begin{figure*}[!t]
  \centering
  \includegraphics[width=\textwidth]{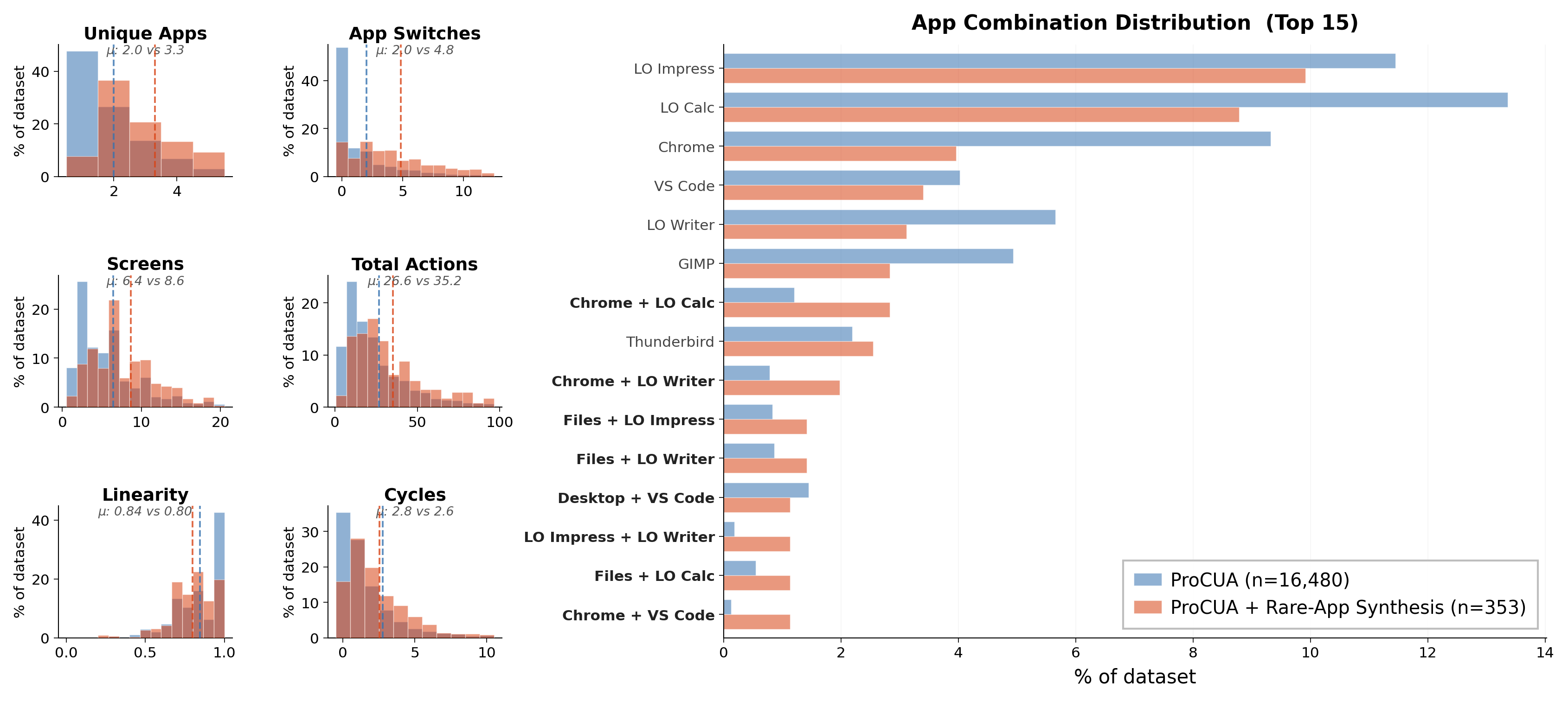}
  \caption{%
    Comparison of the full \textsc{ProCUA} training set
    (\num{16480} trajectories) and the \textsc{ProCUA\,+\,Rare-App Synthesis}
    complexity subset (353~trajectories).
    \emph{Left:} Overlaid histograms (percentage-normalized) for six
    topology metrics; dashed lines mark per-dataset means.
    \emph{Right:} App-combination distribution (top~15 by
    \textsc{ProCUA\,+\,Rare-App Synthesis} frequency).
    Multi-app combinations are shown in bold.
    The rare-app subset shifts toward more applications per trajectory
    (mean 3.3 vs.\ 2.0), more app switches (4.8 vs.\ 2.0),
    more screens (8.6 vs.\ 6.4), and more total actions
    (35.2 vs.\ 26.6), while maintaining comparable cycle counts and only
    moderately lower linearity (0.80 vs.\ 0.84).
    The right panel shows that single-app categories still appear in
    \textsc{ProCUA\,+\,Rare-App Synthesis} but at reduced frequency,
    while multi-app combinations (e.g., Chrome\,+\,LO~Calc,
    Files\,+\,LO~Writer, Desktop\,+\,VS~Code) emerge as a substantial
    fraction of the subset.
  }
  \label{fig:dataset-comparison}
\end{figure*}

% ----- Analysis section -----
\subsection{Complexity-Aware Data Curation}
\label{sec:dataset-comparison}

A key challenge in training GUI agents is the long-tailed distribution of
application combinations: common single-app tasks (e.g., editing a spreadsheet)
dominate the training corpus, while complex multi-app workflows that require
inter-application coordination are rare.
We address this through a two-stage pipeline: (1)~identifying rare,
structurally complex trajectories and using them to synthesize new training
data, and (2)~selecting diverse subsets via round-robin sampling.

\paragraph{Rare-app identification and synthesis.}
Using the app-combination labels extracted during profiling
(\S\ref{sec:trajectory-graph-analysis}), we identify 2{,}065 trajectories
whose app combinations occur $\leq$3 times across the \num{111862}-trajectory
pool---the long tail of rare multi-app workflows.
We then use the goals and trajectory summaries of these rare examples as
few-shot demonstrations in our task synthesis pipeline to generate new
training tasks that target underrepresented multi-app coordination patterns.
The resulting \textsc{ProCUA\,+\,Rare-App Synthesis} subset (353~trajectories) is collected
by executing the synthesized tasks and profiled with full topology metrics.

\paragraph{Topology metric shifts.}
Figure~\ref{fig:dataset-comparison} compares the full \textsc{ProCUA} set and
the rare-app subset across six topology metrics.
The subset exhibits consistently higher structural complexity:
the mean number of unique applications increases from 2.0 to 3.3
(median: 2\,$\to$\,3), and mean app switches more than double from 2.0
to 4.8 (median: 0\,$\to$\,4).
Trajectories visit more screens (8.6 vs.\ 6.4) and execute more GUI actions
(35.2 vs.\ 26.6, median: 29 vs.\ 18).
The mean linearity score decreases only modestly from 0.84 to 0.80, indicating
that even complex multi-app workflows often follow a largely sequential pattern.
Cycle counts are comparable (2.8 vs.\ 2.6), suggesting that navigation loops
arise from task structure (error recovery, data verification) rather than from
application count.

\paragraph{App-combination distribution.}
The full training set is dominated by single-app trajectories: LO~Impress
(13.3\%), LO~Calc (10.5\%), and Chrome (9.7\%) together account for over
one-third of all data.
In the rare-app subset, these categories appear at reduced frequency (8.2\%,
6.2\%, 3.1\%), while multi-app combinations emerge: Chrome\,+\,LO~Calc,
Files\,+\,LO~Writer, Desktop\,+\,VS~Code, and three-app workflows such as
Chrome\,+\,Files\,+\,Terminal each represent 0.6--2.0\% of the subset but
$<$0.1\% of the full set.

\begin{comment}
\paragraph{Diverse subset selection.}
To construct balanced training mixtures from the full pool, we apply a
round-robin selection strategy over application-combination buckets.
Trajectories are grouped by their app combination (2{,}484 unique buckets),
and selection cycles through buckets in order of ascending frequency (rarest
first), drawing one trajectory per bucket per pass.
This guarantees that even singleton app combinations are represented in the
first pass, while common categories are progressively filled in subsequent
rounds.
At $K$\,=\,5{,}000, the selected subset already covers all 2{,}484 app
combinations; at $K$\,=\,10{,}000, it additionally spans 548 of 1{,}007
unique action sets observed in the pool.
\end{comment}

\section{Conclusion}
We introduced ProCUA-SFT, an open dataset of 3.1M SFT samples from 93K fully synthetic computer-use trajectories.
On OSWorld, ProCUA-SFT outperforms human demonstrations for
computer-use SFT\@.
One epoch on UI-TARS~7B reaches 45.0\%---+18.7~pp over the base model and +35~pp over AgentNet's 22.5K human trajectories.
Three design choices prove critical: grounded task synthesis with in-loop
precondition verification ensures that generated goals are feasible in the
current desktop state; seeding desktops with real-world spreadsheets and
presentations produces tasks requiring genuine cross-application reasoning;
and using a single VLM for goal generation, verification, and execution
closes the planner--actor capability gap.
%Diversity-aware curation over 2,484 app-combination buckets further ensures that rare multi-app workflows are represented in the final 3.1M training samples.
%The pipeline runs on commodity KVM instances and serverless GPU inference, with no proprietary desktop environment.
We plan to iterate on ProCUA-SFT as stronger open-weight VLMs, additional
OS platforms, and external reward models become available.

\bibliographystyle{plainnat}
\bibliography{ref}

\end{document}